\documentclass[journal]{IEEEtran}

\usepackage[utf8]{inputenc} 
\usepackage[T1]{fontenc}    
\usepackage{microtype}      

\usepackage{amsmath}        
\usepackage{amsfonts}       
\usepackage{nicefrac}       

\usepackage{graphicx}
\graphicspath{{figures/}}
\usepackage{wrapfig}
\usepackage{float}

\usepackage{booktabs}       
\usepackage{colortbl}
\usepackage{multirow}
\usepackage{tabularx}       
\usepackage{makecell}
\usepackage{ragged2e}       

\usepackage{listings}

\usepackage{xcolor}         
\usepackage{rotating}
\usepackage{url}            
\usepackage{algorithm}
\usepackage{graphicx}
\usepackage{subcaption} 

\renewcommand{\thetable}{\Roman{table}} 
\usepackage{makecell}   
\usepackage{cite}       
\usepackage{amsmath}    

\usepackage{hyperref}       

\newcommand{\customcaption}[1]{%
  {\fontsize{7}{9}\selectfont #1}%
}

\IEEEoverridecommandlockouts

\begin{document}

\title{Lightweight Quad Bayer HybridEVS Demosaicing via State Space Augmented Cross-Attention}

\author{\IEEEauthorblockN{Shiyang Zhou, Haijin Zeng, Yunfan Lu, Yongyong Chen, Jie Liu, \IEEEmembership{Fellow, IEEE}, Jingyong Su}
\thanks{Shiyang Zhou, Yongyong Chen, Jie Liu, and Jingyong Su are with the School of Computer Science and Technology, Harbin Institute of Techonology (Shenzhen), Shenzhen 518055, China. (e-mail: shiyangzhou@stu.hit.edu.cn; cyy2020@hit.edu.cn; jieliu@hit.edu.cn; sujingyong@hit.edu.cn

Haijin Zeng is with the Department of Psychiatry, Harvard University, Cambridge, MA 02138, USA.(e-mail: haijin.zeng2018@gmail.com)

Yunfan Lu is with the AI Thrust, HKUST(GZ) (e-mail: ylu066@connect.hkust-gz.edu.cn)
}
}


\maketitle

\begin{abstract}
Event cameras like the Hybrid Event-based Vision Sensor (HybridEVS) camera capture brightness changes as asynchronous "events" instead of frames, offering advanced application on mobile photography.
However, challenges arise from combining a Quad Bayer Color Filter Array (CFA) sensor with event pixels lacking color information, resulting in aliasing and artifacts on the demosaicing process before downstream application.
Current methods struggle to address these issues, especially on resource-limited mobile devices.
In response, we introduce \textbf{TSANet}, a lightweight \textbf{T}wo-stage network via \textbf{S}tate space augmented cross-\textbf{A}ttention, which can handle event pixels inpainting and demosaicing separately, leveraging the benefits of dividing complex tasks into manageable subtasks.
Furthermore, we introduce a lightweight Cross-Swin State Block that uniquely utilizes positional prior for demosaicing and enhances global dependencies through the state space model with linear complexity.
In summary, TSANet demonstrates excellent demosaicing performance on both simulated and real data of HybridEVS while maintaining a lightweight model, averaging better results than the previous state-of-the-art method DemosaicFormer across seven diverse datasets in both PSNR and SSIM, while respectively reducing parameter and computation costs by $1.86\times$ and $3.29\times$.
Our approach presents new possibilities for efficient image demosaicing on mobile devices. Code is available in the supplementary materials.
\end{abstract}


\IEEEpeerreviewmaketitle

\section{Introduction}
\label{sec:intro}

\begin{figure}[t]
	\centering
    \subfloat{\includegraphics[width=0.1\textwidth]{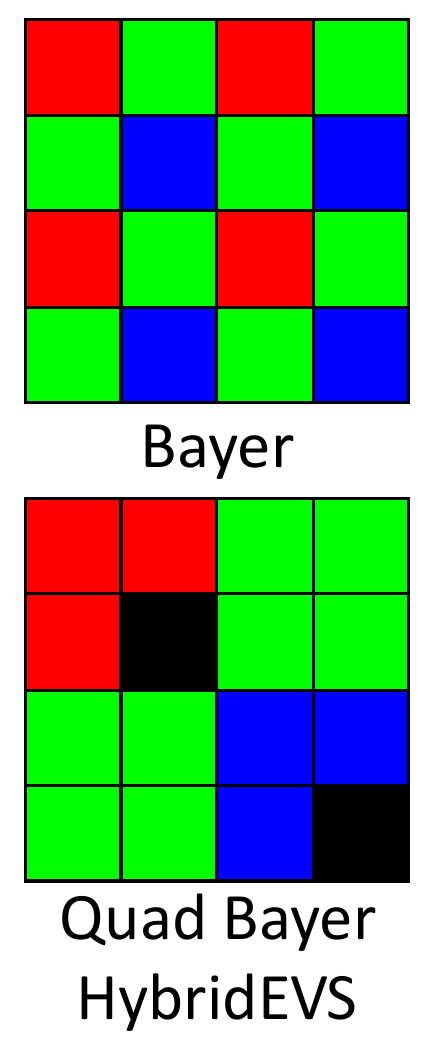}\label{fig: event}}
    \hfill
    \subfloat{\includegraphics[width=0.36\textwidth]{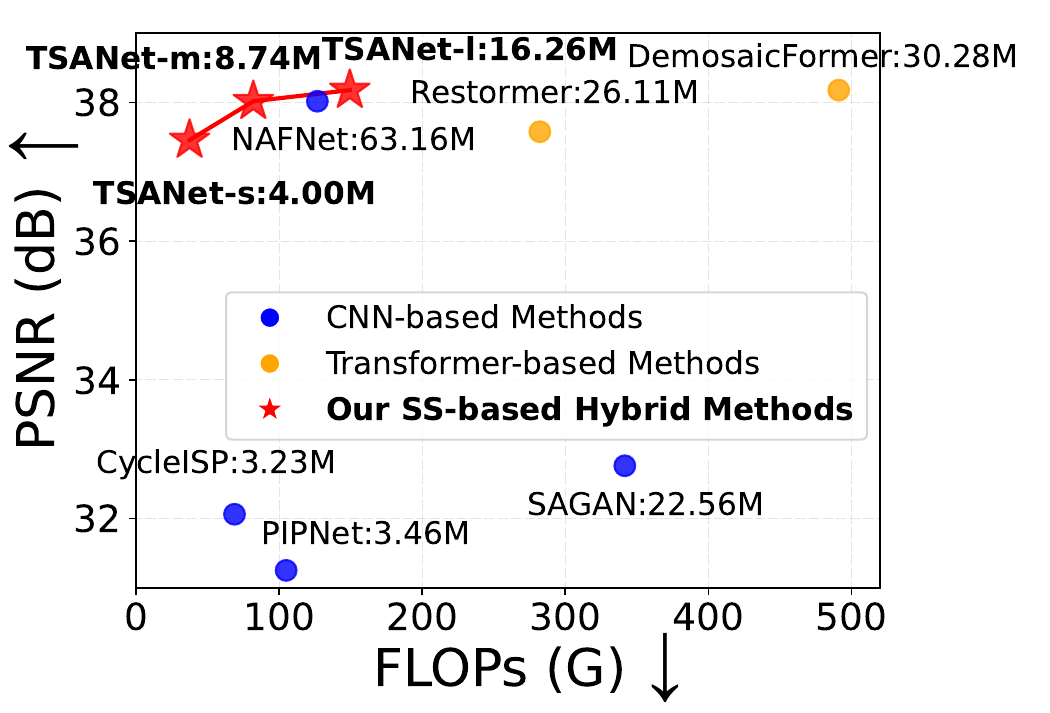}\label{fig: comparison}}
	\caption{\small Left: Bayer CFA vs. Quad Bayer CFA on HybridEVS. Right: PSNR vs. FLOPs running on seven datasets, each model indicates its respective parameters on the right side. Our state space (SS) based TSANet presents three different sizes, each delivering optimal performance across various complexity ranges.}
	\label{figure1}
    \vspace{-5mm}
\end{figure}

In recent years, event cameras have made significant progress as a new type of image sensor. Compared to conventional digital cameras, event cameras can capture event information by sensing the intensity change of specialized event pixels, thereby capturing information about objects' movement~\cite{litzenberger2006embedded,4444573}. However, basic imaging before downstream applications is essential for event cameras, which have not received adequate attention. For the critical demosaicing process in imaging, it has been found that traditional Bayer CFA sensors are constrained by their design on mobile devices, making it difficult to attain high-quality images in low-light scenes~\cite{kim2019high}. Therefore, non-Bayer CFA sensors have become mainstream in mobile photography in recent years, utilizing specifically designed CFA to enhance low-light imaging performance~\cite{okawa20191}. Quad Bayer is one of the most popular formats, which can acquire high-resolution images on common scenes and enhance low-light imaging by pixel-binning~\cite{yoo2015low}. Nevertheless, different from Bayer CFA which has been proposed over several decades~\cite{maschal2010review,7567576}, exploration of Quad Bayer CFA is still very limited~\cite{yang2022mipi}.
A type of event camera design named Hybrid Event Vision Sensors~\cite{kodama20231} involves utilizing a Quad Bayer CFA on a camera sensor and allocating certain pixels as event pixels to capture motion data instead of RGB color information, as shown in Fig. \ref{figure1} left. The non-conventional Quad Bayer arrangement and the absence of color information at event pixel locations pose challenges for the demosaicing process, traditional methods face difficulties in extracting patterns from such complex arrangements, resulting in reduced imaging quality and poor performance in downstream applications such as deblurring and object detection. 
To improve the restoration quality, recent research~\cite{zhou2018deep,kim2019high,wu2024mipi} has proposed some end-to-end demosaicing methods based on convolution neural networks (CNNs) and Transformers. CNN-based methods have shown impressive results on Bayer images but fall short when applied to Non-Bayer images. Transformer-based approaches leverage long-range dependencies for effective demosaicing but are limited by high computational demands for edge device implementation. Furthermore, previous methods have overlooked the strong relationship between the CFA and RAW images, typically addressing the issue only in an end-to-end manner.

Overall, several challenges for demosaicing in Quad Bayer HybridEVS cameras are encountered: \textbf{i)} How to alleviate the decrease in reconstruction quality resulting from the absence of color information at event pixels locations; \textbf{ii)} How to reduce the model's parameter and computational complexity to make it valuable for practical applications in edge computing. 

\begin{figure*}[t]
    \centering
    \includegraphics[width=1\textwidth]{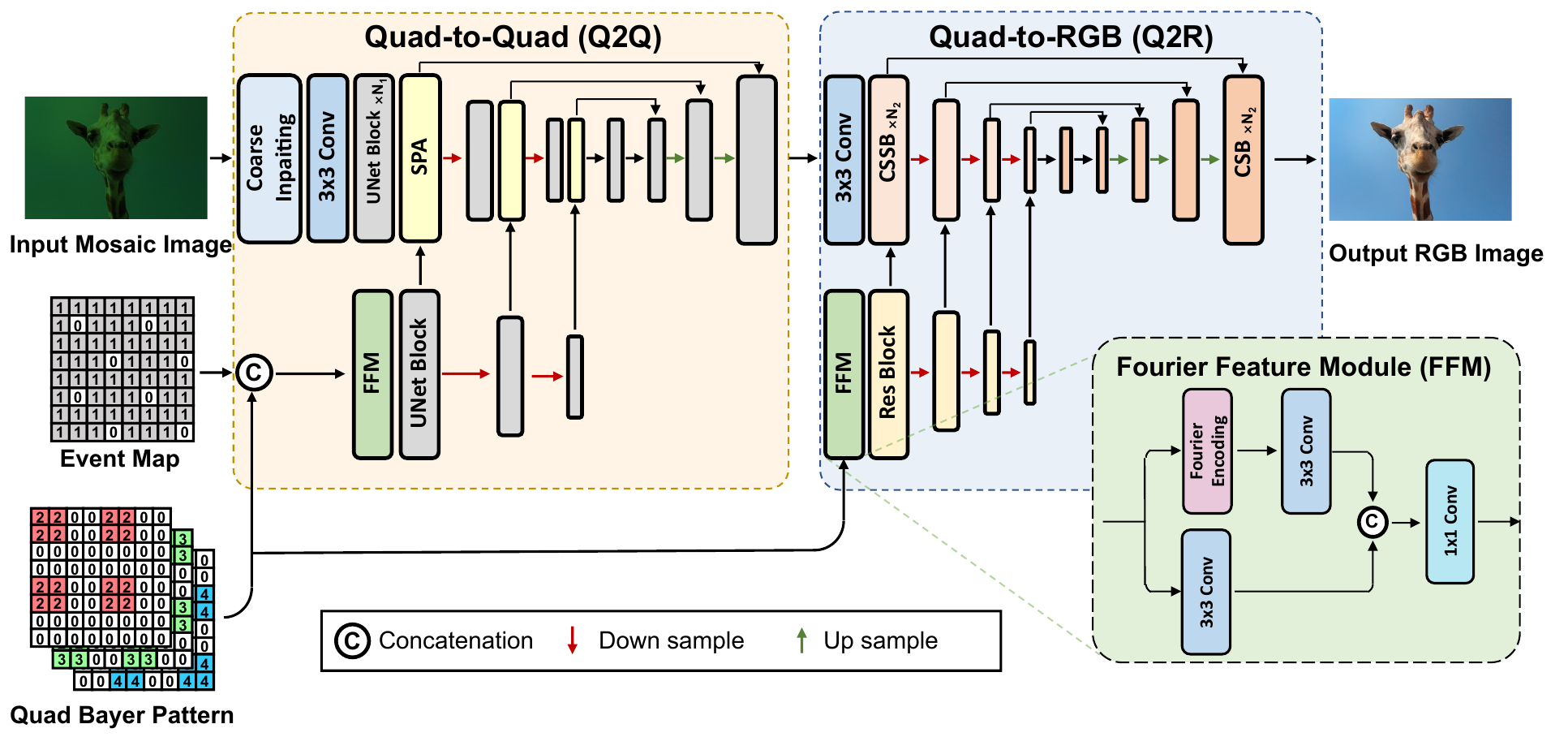}
    \caption{{Overview of the TSANet approach. We adopt a two-stage structure that breaks down complex tasks into manageable subtasks while leveraging additional branches to utilize position prior.}}
    \label{fig:model}
    \vspace{-4mm}
\end{figure*}

To address the two issues, we introduce TSANet, a novel lightweight two-stage model that effectively combines the position information and color information, augmented by the state space model to further explore long-range relationships in linear complexity. Specifically, to improve computational efficiency while enhancing reconstruction quality, we divide the complex task demosaicing for HybridEVS into two stages, as shown in Fig. \ref{fig:model}. The initial stage, termed Quad-to-Quad (Q2Q), is dedicated to inpainting event pixels while the second Quad-to-RGB (Q2R) stage is tailored for Quad Bayer demosaicing. Both stages employ U-Net-like networks, with an extra position branch for each sub-network. A two-step training strategy is additionally employed to effectively increase stability and robustness ~\cite{zeng2023inheriting}.

For the design of specific models, inspired by~\cite{zheng2024quad,sun2022event}, we employ an extra position branch that introduces an explicit position encoding, enabling the network to have prior knowledge of positional relationships. For both stages, we introduce an attention mechanism to fuse positional prior information (see Sec. \ref{attentions}). Additionally, to capture long-range dependencies and enhance the demosaicing effect, we implement the Residual Vision State Space (RVSS) model~\cite{zhu2024vision} in parallel with local feature extraction. The hybrid design fuses local window attention across position and image, while efficiently incorporating global spatial information through RVSS with linear complexity. 
In summary, our contributions are as follows:
\begin{itemize}
\item We propose TSANet, a lightweight \underline{T}wo-stage network via \underline{S}tate space augmented cross-\underline{A}ttention. By employing a designed sub-tasks allocation and dual-branch encoders, TSANet achieves state-of-the-art performance with fewer computational resources (See Fig. \ref{figure1}).

\item We present two unique state space augmented cross-attention blocks. The combined use of the Residual Vision State Space module with local attention and convolution demonstrates great effectiveness advantages, leading to outstanding capabilities in local-global feature extraction in a linear format.

\item We design two cross-modality attention mechanisms between position and image information, which based on local window attention and dot product, respectively, effectively capture position features, facilitating information exchange and integration across different modalities.

\end{itemize}

\section{Related Work}
\paragraph{Bayer Demosaicing}
Traditional approaches for Bayer demosaicing primarily rely on interpolation techniques~\cite{hirakawa2006joint,6171859}, utilizing methods like adaptive algorithm ~\cite{hirakawa2005adaptive} and spatial-spectral correlations~\cite{li2008image} to reconstruct full-color images. Recently, the success of convolutional networks (CNNs) used in deep learning has led to great progress in demosaicing~\cite{syu2018learning,tan2017color,tan2017joint,tan2018deepdemosaicking,liu2020joint}. These methods replace traditional interpolation techniques with deep neural networks, leveraging the powerful fitting capabilities of neural networks to achieve excellent results. Some researchers~\cite {liu2020joint,9503334,zhang2022deep} proposed a demosacing network to utilize internal image information like color prior and CFA arrangement. However, most of these methods can't be extended to non-Bayer CFA formats, facing limitations like artifacts and aliasing when dealing with more challenging non-Bayer formats like Quad Bayer.

\paragraph{Quad Bayer Demosaicing}
In recent years, Quad Bayer has become a popular CFA pattern widely used in mobile photography to improve low-light imaging performance, such as smartphone cameras~\cite{yang2022mipi,zhou2025binarized}. Different from traditional Bayer CFA, exploration of the Quad Bayer CFA is limited. The larger gaps between pixels of the same color make the task more challenging compared to Bayer CFA. Some two-stage networks~\cite{jia2022learning,zeng2023inheriting} are proposed for progress learning to enhance Quad Bayer demosaicing. Zhang \emph{et al.}\cite{zheng2024quad} proposed a dual-encoder structure to achieve better joint demosaicing and denoising tasks. Xu et al.~\cite{Xu_2024_CVPR} introduced a two-stage transformer architecture that first generates a coarse output and subsequently refines it in the second stage for superior detail recovery. GAN-based networks~\cite{a2021beyond,sharif2021sagan} are also used to strengthen the high-frequency details of RGB images for Non-Bayer CFA. The sensor's CFA offers a strong positional prior, yet most methods overlook the valuable color arrangement information, resulting in significant color distortion in Quad Bayer HybridEVS demosaicing.

\paragraph{Event Camera Imaging}
As a novel type of sensor that has emerged in recent years~\cite{gallego2020event,son2017640}, the imaging research of event cameras is an active topic. Recent studies have mainly focused on imaging or downstream applications like object detection~\cite{zhang2022spiking} based on event information. Scheerlinck \emph{et al.}\cite{scheerlinck2020fast} introduced a fast image restoration method with deep neural networks. These methods have demonstrated excellent performance in visual tasks based on event information. However, for the latest proposed hybrid event-based vision sensor~\cite{kodama20231}, the field is almost blank in advanced methods for its Image Sensor Pipelines (ISPs). A crucial step in ISPs is the demosaicing process, which converts Quad Bayer data into the RGB domain and directly impacts the imaging quality for downstream applications.

\paragraph{State Space Models}
In recent years, State Space Models (SSMs)~\cite{gu2021efficiently,} have emerged as competitive rivals to traditional deep learning architectures like Convolution Neural Networks (CNNs) and Transformers. Pioneering works like S4~\cite{gu2021efficiently} and S5~\cite{smith2022simplified} introduced advancements on deep-state models with efficient parallel scan, modeling long-range dependency. Recently proposed Mamba~\cite{gu2023mamba}, featuring a data-dependent SSM layer, has shown remarkable performance, surpassing Transformers in natural language tasks with linear scalability in sequence length. Additionally, some works have applied Mamba to various vision tasks and multimodel tasks, including image classification~\cite{zhu2024vision}, video understanding~\cite{wang2023selective,ren2025vamba}, image restoration~\cite{guo2024mambair,10891450,10817590,10604894,guo2025mambairv2}, segmentation~\cite{liao2024lightm, 10891410}, multimodel understanding ~\cite{li2025alignmamba,zhang2025mamba}, demonstrates the potential in visual tasks for lightweight models.

\section{Methods}
Initially, we present the overall pipeline of our two-stage model in Sec. \ref{two_stage}. Then we introduce the proposed state space augmented-cross attention block and its variant in Sec. \ref{ssablocks}. Finally, we discuss details of our proposed attention mechanisms across position and image in Sec. \ref{attentions}. 

\subsection{Two-stage Network Strcture}\label{two_stage}

For the task of Quad Bayer HybridEVS demosaicing, our goal is to recover a three-channel RGB image $\mathbf{I_R} \in \mathbb{R}^{H\times W\times 3}$ from a degraded Quad Bayer image $\mathbf{I_Q} \in \mathbb{R}^{H\times W\times 1}$, where the degradation includes color channel loss $D_q$ due to the Quad Bayer CFA, pixel absence $D_e$ due to the design of event pixels in the sensor (see Fig. \ref{figure1}), respectively. Most previous methods aim to directly learn the entire process through an all-in-one deep model $\mathcal{M}$ that restores image $\mathbf{I_R}$ from $\mathbf{I_Q}$, which can be expressed as:
\begin{equation}
\mathbf{I_R} = \mathcal{M}(\mathbf{I_Q}).
\end{equation}
However, these all-in-one models often struggle to extract the inner connection between position and color, causing unbearable aliasing and artifacts (See Fig. \ref{figure6}), or require a large number of parameters and computation load to achieve ideal restoration results, making it barely impossible to deploy on limited-resource mobile devices. Unlike past single-model solutions, we define the composite task as two controllable sub-tasks: the former sub-task is to restore the degradation of $D_e$ and $D_n$ from the original Quad Bayer image~\cite{8820082}, inpainting absent pixels, producing a clean Quad Bayer image, defined as $\mathcal{M}^{Q2Q}$; the later one is to restore the clean Quad Bayer image into $\mathbf{I_R}$, defined as $\mathcal{M}^{Q2R}$. The overall pipeline progressively restores the RGB image from the degraded Quad Bayer image. Notably, we introduce a distinctive encoder branch designed to integrate position information as a dedicated prior knowledge into the network, using position information, event map $\mathbf{P_e}\in \mathbb{R}^{H\times W\times 1}$ from $D_e$ and Quad Bayer CFA map $\mathbf{P_q}\in \mathbb{R}^{H\times W\times 3}$ from $D_q$ 
 (see Fig.\ref{fig:model}) to achieve better image reconstruction, the process can be expressed as:
\begin{equation}
\mathbf{I_R} = \mathcal{M}^{Q2R}(\mathcal{M}^{Q2Q}(\mathbf{I_Q}, (\mathbf{P_q}, \mathbf{P_e})),\mathbf{P_q}).
\end{equation}
Our network adopts a two-stage strategy on both model design and training recipe, which demonstrates a dedicated design for the complicated demosaicing task on HybridEVS. The two-stage network architecture is depicted in Fig. \ref{fig:model}. Such a design not only assigns specific tasks to sub-networks but also benefits from a two-step training strategy. Prior studies~\cite{zeng2023inheriting, Xu_2024_CVPR} have demonstrated that pretraining on sub-networks can lead to improved performance and inference stability. The Q2Q and Q2R networks are first trained separately on their respective tasks, and then fine-tuned jointly. Our designed two-stage architecture facilitates directional pretraining for the two sub-networks by synthesizing a dummy clean Quad Bayer (see details in Sec.\ref{details}). 

Besides, we believe fully utilizing position information is crucial for the demosaicing problem of Quad Bayer, especially in the task with event pixels. Therefore, we propose an additional positional encoding branch in both networks, explicitly integrating position information into the network. To further enhance the model's ability to extract high-frequency texture information, we designed a \textbf{F}ourier \textbf{F}eature \textbf{M}odule (FFM) based on Fourier encoding, as shown in Fig. \ref{fig:model}, which maps the position information to a series of high-frequency features based on sine and cosine functions~\cite{tancik2020fourier}, enabling the model to capture refine details and patterns in position dimension, strengthening the restoration results of complex textures. At the Q2Q stage, before putting the Quad Bayer image into the network, we apply a coarse inpainting by averaging nearby pixels around event pixels, which aims to mitigate color loss resulting from event pixels.

\subsection{State Space Augmented Blocks}\label{ssablocks}

In this section, we propose two lightweight modules augmented by State Space Models for the encoder and decoder of the Q2R stage, respectively. We begin with a dual-branch structure named Cross-Swin State Block (CSSB), concurrently modeling local cross-modality attention and long-range dependencies with linear complexity. The design also exploits Quad Bayer and event patterns as strong structural priors to improve demosaicing. Then, we present its variant Conv State Block (CSB) in the decoder, enhancing local feature restoration while further reducing computation by convolutions. 

\begin{figure}[t]
    \centering
    \subfloat[Cross-Swin State Block]{\includegraphics[width=0.47\textwidth]{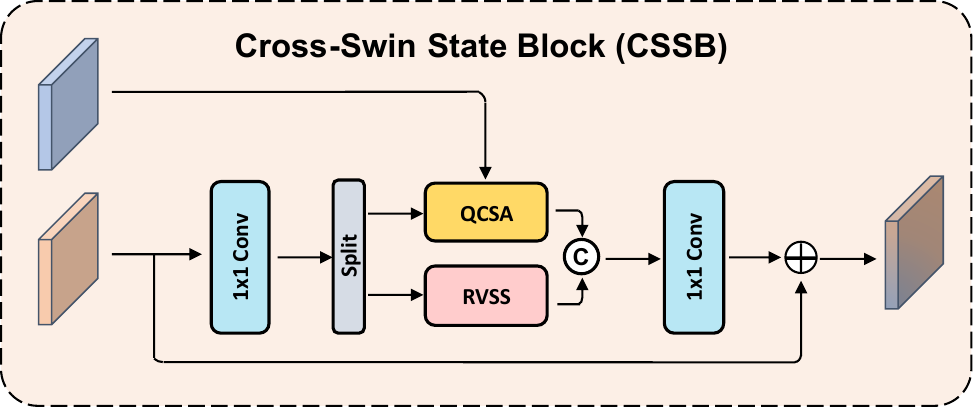}\label{fig: cssb}}
    \vfill
    \subfloat[Conv State Block]{\includegraphics[width=0.47\textwidth]{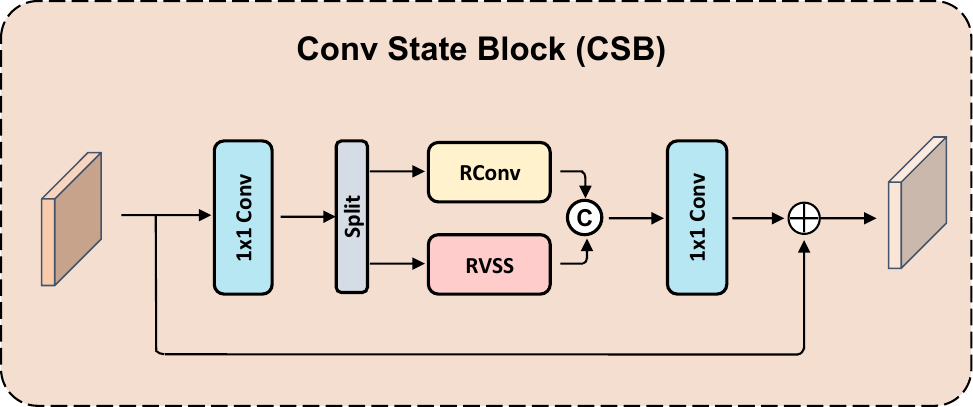}\label{fig: csb}}
    \caption{Proposed state space augmented blocks. The two blocks are modified from Transformer and Convolution for the Q2R encoder and decoder, respectively. The dual-branch design parallel extracts feature with local-global dependencies while reducing computation load.}
    \label{fig: figure3}
    
    \vspace{-1mm}
\end{figure}

\paragraph{Cross-Swin State Block}\label{sec_cssb}
Fig. \ref{fig: cssb} illustrated our proposed Cross-Swin State Block (CSSB). This block is designed to capture long-range dependencies and cross-modality local attention in parallel. It integrates Residual Vision State Space (RVSS) and Quad Bayer Cross Swin Attention (QCSA) mentioned in Sec. \ref{qcsa}. Fig. \ref{fig: rvss} shows RVSS, a simplified version of Residual State Space Block of MambaIR~\cite{guo2024mambair}, preserving its core component VSSM for efficient extraction of long-range dependencies, followed by a residual connection. To further reduce computational complexity while capturing local positional attention intersections and global long-range dependencies simultaneously, we follow SCUNet~\cite{zhang2023practical} and propose a parallel network module. First, the image feature projected through 1x1 convolution is split and separately inputted into QCSA and RVSS modules. The outputs are then concatenated and utilized for out projection through a $1\times 1$ convolution, which is followed by a residual connection. For a image input $\mathbf{F_I}$ and a position input $\mathbf{F_P}$, the process can be expressed as follows:
\begin{equation}
\begin{split}
    {X_1}, {X_2} &= \operatorname{Split}(\operatorname{Conv_{1\times1}}(\mathbf{F_I})),\\
    {Y_1}, {Y_2} &= \operatorname{QCSA}(X_1, \mathbf{F_P}),\  \operatorname{RVSS}(X_2),\\
    \hat{\mathbf{F}}_{\mathbf{I}} &= \operatorname{Conv_{1\times1}}(\operatorname{Concat}(Y_1, Y_2))+\mathbf{F_I}.
\end{split}
\end{equation}
The QCSA primarily extracts representations with position and spatial information through the two modalities of position and image while RVSS is used to effectively capture global information, which parallel addresses cross-modality local attention and long-range dependencies. This parallel design allows the model to independently extract two distinct types of features--local high-frequency textures and global low-frequency structures. Moreover, the dual-brunch structure is similar to group convolution~\cite{zhang2017interleaved}, reducing the number of channels within the module through splitting operations, effectively reducing the computational complexity and parameters of the block. RVSS employs multi-directional linear selective scanning to process pixel sequences, dynamically focusing via context retention and control matrices. Multi-directions (4 in ours) ensure global attention, maintaining color/texture consistency while reducing aliasing, with selective long-range integration addressing lightweight challenges for edge devices like HybridEVS.

\paragraph{Conv State Block}
We also propose a variant of CSSB for the decoder of the Q2R stage, as shown in Fig. \ref{fig: csb}. Instead of employing QCSA to capture attention features across position information, we replace this module with Residual Convolution (RConv)~\cite{zhang2021plug}, which focuses on restoring internal local feature of the image, parallel with an RVSS to enhance long-range feature extraction capability, forming a lightweight decoder block with short-long dependencies.

\subsection{Attention Modules for Position Fusion}\label{attentions}
\begin{figure*}[t]
    \centering
    \subfloat[Quad Bayer Cross Swin Attention]{\includegraphics[width=0.33\textwidth]{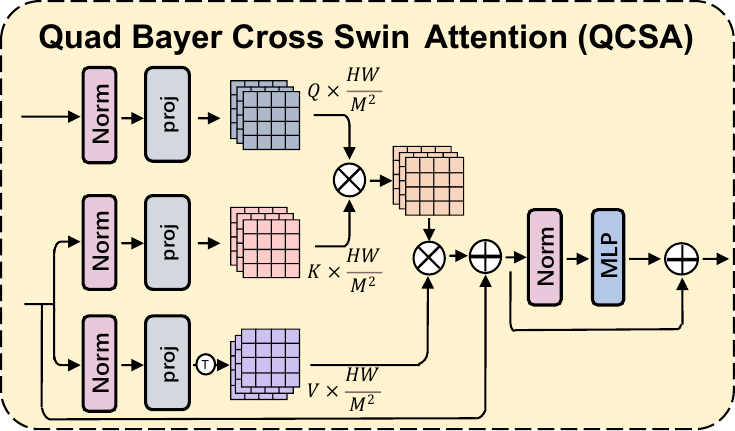}\label{fig: qcsa}}
    \hfill
    \subfloat[Spatial Position Attention]{\includegraphics[width=0.33\textwidth]{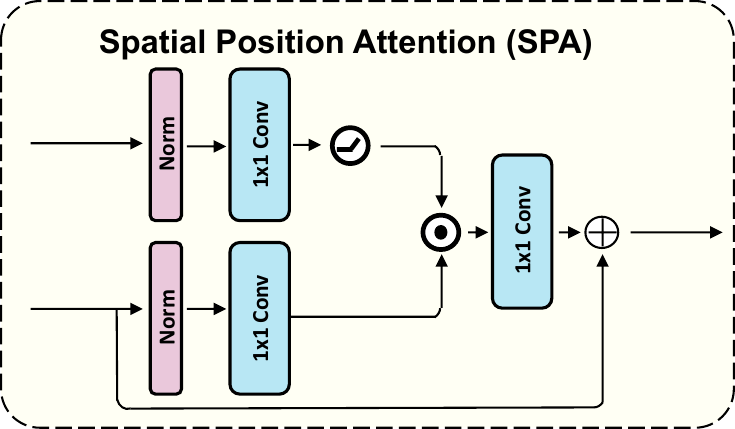}\label{fig: psa}}
    \hfill
    \subfloat[Residual Vision State Space]{\includegraphics[width=0.33\textwidth]{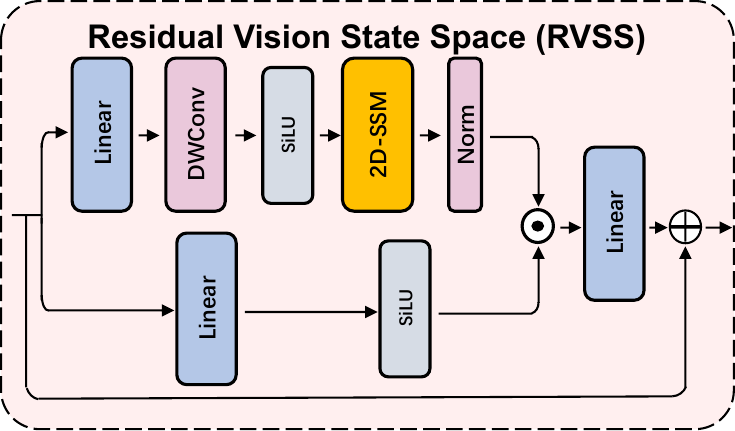}\label{fig: rvss}}
    \caption{Illustration of designed modules. We propose two cross-attention modules and one state space model. Attentions focus on fusing position information into the network and RVSS captures long-range dependencies in linear complexity.}
    \label{fig: figure4}
    \vspace{-3mm}
\end{figure*}

As illustrated in Fig. \ref{fig: qcsa} and  Fig. \ref{fig: psa}, we propose two spatial attention modules designed for fusing position information and image features. We utilize the position representations as extra information and explore the spatial relationship of position and image.

\paragraph{Quad Bayer Cross Swin Attention}\label{qcsa}
As shown in Fig. \ref{fig: qcsa}, we designed a sufficient feature fusion mechanism for the Q2R sub-network. It first employs an efficient Window Multi-head Cross Attention (WMCA) inspired by Swin Transformer~\cite{liu2021swin}. Specifically, given image feature $\mathbf{F_I} \in \mathbb{R}^{H\times W\times d}$ and position feature $\mathbf{F_P} \in \mathbb{R}^{H\times W\times d}$, both features are first partitioned into non-overlapping $M\times M$ local windows, getting $\frac{HW}{M^2}\times{M^2}\times{d}$ features. Different from conventional attention acquiring all $query (Q)$, $key (K)$, $value (V)$ from $\mathbf{F_I}$, we produce $Q$ from $\mathbf{F_P}$. The $Q$, $K$ and $V$ for window feature $X \in \mathbb{R}^{{M^2}\times d}$ from $\mathbf{F_I}$ and $Y \in \mathbb{R}^{{M^2}\times d}$ from $\mathbf{F_P}$ are computed as:
\begin{equation}
Q = Y{P_Q},\quad K=X{P_K},\quad V=X{P_V},
\end{equation}
where ${P_Q}$, ${P_K}$ and ${P_V}$ are shared project matrices among each window. Then we have $Q, K, V \in \mathbb{R}^{{M^2}\times d}$ to  compute in-window cross attention as:
\begin{equation}
\text { Attention }(Q, K, V)=\operatorname{SoftMax}\left(\frac{Q K^T }{\sqrt{d}}+B\right),
\end{equation}
where $B$ represents relative positional encoding within the window, complementing the global positional encoding introduced by $\mathbf{F_P}$. This attention mechanism operates across all windows and is executed in parallel $h$ times~\cite{vaswani2017attention}. After WMCA, the features are then fed into a two-layer multi-layer perceptron (MLP) with GELU~\cite{hendrycks2016gaussian} activation function for further feature extraction. Both steps utilize residual connections and LayerNorm~\cite{ba2016layer}, the whole process can be expressed as:
\begin{equation}
    \begin{split}
        \hat{\mathbf{F}}_{\mathbf{I}}&=\textrm{WMCA}(\operatorname{LN}(\mathbf{F_I}),\operatorname{LN}(\mathbf{F_P}))+\mathbf{F_I},\\
        \hat{\mathbf{F}}_{\mathbf{I}}&=\textrm{MLP}(\hat{\mathbf{F}}_{\mathbf{I}})+{\hat{\mathbf{F}}_{\mathbf{I}}}.
     \end{split}
\end{equation}
Then, through the shifted window mechanism~\cite{liu2021swin}, this module achieves cross-window information exchange. It incorporates global information of Quad Bayer CFA pattern into the network, offering a spatial attention mechanism based on position information. Additionally, the window mechanism maintains the computational complexity of the network linearly, effectively controlling computational and parameter overhead compared to conventional attention methods. This module achieves efficient position-image interaction via shifted-window local attention with linear complexity, which is critical for performance-deployability balance.

\paragraph{Spatial Position Attention}\label{sec_spa}
As shown in Fig. \ref{fig: psa}, we designed another cross-modality attention mechanism, aimed at building the relationship between position and image sufficiently in the Q2Q stage. Specifically, the position branch introduces an explicit representation of the spatial dimension. Firstly, the position feature $\mathbf{F_P} \in \mathbb{R}^{H\times W\times d}$ and the image feature $\mathbf{F_I} \in \mathbb{R}^{H\times W\times d}$ pass through their respective convolutional projection layers. Subsequently, the position branch is activated by ReLU, after which they demonstrate element-wise product with each other. The process can be expressed as:
\begin{multline}
\hat{\mathbf{F}}_{\mathbf{I}} = \operatorname{Conv}_{1 \times 1}\left(\operatorname{LN} \left( {\mathbf{F}}_{\mathbf{I}} \right)\right) \odot \\
\operatorname{ReLU}(\operatorname{Conv}_{1 \times 1}(\operatorname{LN} \left( {\mathbf{F}}_{\mathbf{P}} )\right)) + {\mathbf{F}}_{\mathbf{I}}.
\end{multline}
The position information is calculated to a weights map of the image feature, determining the importance of each feature pixel. It is similar to a Gate Mechanism~\cite{zamir2022restormer}, which can control the flow of information but by position information rather than itself. Additionally, this method exhibits faster computational performance when compared to convolution or attention mechanisms.

\begin{figure*}
    \fontsize{10}{8}\selectfont
         \vfill
     \begin{subfigure}[b]{0.30\textwidth}
         \centering
         \includegraphics[width=\linewidth, trim=20 10 21 0, clip]{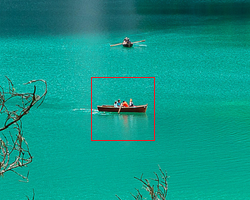}
         \caption*{\customcaption{0833, Test Set\\ MIPI~\cite{wu2024mipi}}}
         \label{fig: f1ref}
     \end{subfigure}
     \hfill
     \begin{subfigure}[b]{0.13\textwidth}
         \begin{subfigure}[b]{\textwidth}
             \centering
             \includegraphics[width=\linewidth]{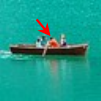}
             \caption*{\customcaption{{\makecell{Reference\\PSNR/Parameters}}}}
             \label{fig: f3gt}
             \end{subfigure}
         \begin{subfigure}[b]{\textwidth}
             \centering
             \includegraphics[width=\linewidth]{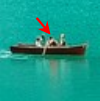}
             \caption*{\customcaption{\makecell{DemosaicFormer~\cite{Xu_2024_CVPR}\\26.97dB/30.28M}}}
             \label{fig: f3restormer}
             \end{subfigure}
     \end{subfigure}
     \hfill
     \begin{subfigure}[b]{0.13\textwidth}
        \begin{subfigure}[b]{\textwidth}
             \centering
             \includegraphics[width=\linewidth]{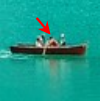}
             \caption*{\customcaption{\makecell{NAFNet~\cite{chen2022simple}\\26.41dB/63.16M}}}
             \label{fig: f3nafnet}
         \end{subfigure}
         \begin{subfigure}[b]{\textwidth}
             \centering
             \includegraphics[width=\linewidth]{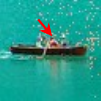}
             \caption*{\customcaption{\makecell{SAGAN~\cite{sharif2021sagan}\\18.89dB/22.56M}}}
             \label{fig: f3sagan}
         \end{subfigure}
     \end{subfigure}
     \hfill
     \begin{subfigure}[b]{0.13\textwidth}
         \begin{subfigure}[b]{\textwidth}
             \centering
             \includegraphics[width=\linewidth]{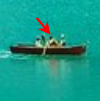}
             \caption*{\customcaption{\makecell{SwinIR~\cite{liang2021swinir}\\ 25.02dB/2.86M}}}
             \label{fig: f3pipnet}
             \end{subfigure}
         \begin{subfigure}[b]{\textwidth}
             \centering
             \includegraphics[width=\linewidth]{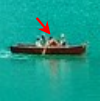}
             \caption*{\customcaption{\makecell{UFormer~\cite{wang2022uformer}\\ 26.21dB/14.04M}}}
             \label{fig: f1gt}
         \end{subfigure}
     \end{subfigure}
     \hfill
     \begin{subfigure}[b]{0.13\textwidth}
         \begin{subfigure}[b]{\textwidth}
             \centering
             \includegraphics[width=\linewidth]{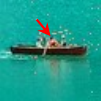}
             \caption*{\customcaption{\makecell{PIPNet~\cite{a2021beyond}\\22.78dB/3.46M}}}
             \label{fig: f3pipnet}
             \end{subfigure}
         \begin{subfigure}[b]{\textwidth}
             \centering
             \includegraphics[width=\linewidth]{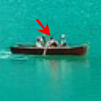}
             \caption*{\customcaption{\makecell{Restormer~\cite{zamir2022restormer}\\ 25.74dB/26.11M}}}

             \label{fig: f1gt}
         \end{subfigure}
     \end{subfigure}
     \label{fig: fig3}
     \hfill
     \begin{subfigure}[b]{0.13\textwidth}
         
            \begin{subfigure}[b]{\textwidth}
             \centering
             \includegraphics[width=\linewidth]{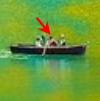}
             \caption*{\customcaption{\makecell{CycleISP~\cite{zamir2020cycleisp}\\10.25dB/3.23M}}}
             \label{fig: f3pipnet}
             \end{subfigure}
         \begin{subfigure}[b]{\textwidth}
             \centering
             \includegraphics[width=\linewidth]{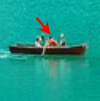}
             
             \caption*{\customcaption{\makecell{{\textbf{TSANet-l} (Ours)}\\ \textbf{27.04dB}/16.26M}}}

             \label{fig: f1gt}
         \end{subfigure}
     \end{subfigure}
     
    \caption{Visualized results of all compared methods for Quad Bayer HybridEVS Demosaicing on MIPI dataset. The proposed TSANet demonstrates the best visual results among all methods, producing more vivid colors on scenes with complex coloration, better than the previous state-of-the-art approach DemosaicFormer.}

\label{figure5}
\end{figure*}
\section{Experiments}
In this section, we first introduce implementation details about both training and testing. Then we illustrate all the datasets we use, including synthesized and real Quad Bayer HybridEVS data. Then we compared our proposed TSANet with other state-of-the-art methods on both real HybridEVS data and seven diverse synthesized datasets. Finally, we demonstrate an ablation study to prove the effectiveness of our methods.

\subsection{Experimental settings}

\paragraph{Implementation details}\label{details}
In all experiments, we use the following hyperparameters, unless mentioned otherwise. During the training, we randomly crop the Quad Bayer input and RGB ground truth into $128\times128$ patches, with batch size = 32. We use Adam as the optimizer and the learning rate starts from $2\times10^{-4}$ and is reduced to $1\times10^{-7}$ with the cosine annealing scheme. The total iteration is set to $1\times10^6$. We apply Charbonnier loss during training. Specifically, to fully utilize the two-stage structure of our TSANet, we apply a pretraining step on the sub-network before end-to-end joint training. We synthesized clean Quad Bayer images from RGB images to serve as the ground truth for Q2Q and the input for Q2R. In particular, we employ the published pretrained weights of DemosaicFormer for testing. The training process is conducted on 4 RTX 4090 and the testing process is finished on 1 RTX 4090.

\begin{table*}[ht]
  \caption{Quantitative comparison on image datasets. TSANet variants achieve top performance across all five benchmarks with significantly lower complexity.}
  \centering
  \small
  \setlength{\tabcolsep}{1.2mm}{
    \begin{tabular}{l|cc|ccccc|c}
      \toprule
      \multirow{2}{*}{Methods} & \multirow{2}{*}{\makecell{Params\\(M)}} & \multirow{2}{*}{\makecell{FLOPs\\(G)}} & Kodak & BSD100 & Urban100 & Wed & MIPI & \multirow{2}{*}{Average} \\
      \cline{4-8}
      & & & \multicolumn{5}{c|}{PSNR / SSIM} & \\
      \midrule
      \rowcolor{green!5}
      CycleISP~\cite{zamir2020cycleisp} & \textbf{3.23} & 104.9 & 33.09/0.970 & 32.18/0.969 & 29.78/0.942 & 30.22/0.944 & 30.04/0.934 & 31.06/0.952 \\
      \rowcolor{green!5}
      PIPNet~\cite{a2021beyond} & 3.46 & 68.8 & 32.20/0.960 & 31.97/0.950 & 28.92/0.942 & 29.19/0.929 & 33.73/0.950 &  31.20/0.946\\
      \rowcolor{orange!10}
      SwinIR~\cite{liang2021swinir} & 2.86 & 407.2 & 37.96/0.981 & 36.25/0.981 & 35.29/0.976 & 34.82/0.969 & 37.68/0.978 &  36.40/0.977\\
      \rowcolor{orange!10}
      UFormer~\cite{wang2022uformer} & 14.04 & 55.9 & 38.51/0.984 & \textbf{36.98}/0.984 & 35.51/0.977 & 35.01/0.971 & 38.00/0.980 & 36.80/0.979 \\
      \rowcolor{gray!20}
      \textbf{TSANet-s (Ours)} & 4.00 & \textbf{37.4} & \textbf{38.73/0.984} & 36.56/\textbf{0.984} & \textbf{36.15/0.980} & \textbf{35.19/0.973} & \textbf{38.47/0.978} & \textbf{37.02/0.980} \\
      \midrule
      \rowcolor{green!5}
      SAGAN~\cite{sharif2021sagan} & 22.56 & 341.6 & 36.14/0.974 & 30.53/0.931 & 29.89/0.946 & 28.22/0.917 & 34.25/0.959 & 31.81/0.945\\
      \rowcolor{green!5}
      NAFNet~\cite{chen2022simple} & 63.16 & 126.7 & 39.04/0.985 & \textbf{37.51}/0.986 & 36.51/0.982 & \textbf{35.73}/0.969 & 38.89/0.979 &  37.54/0.980\\
      \rowcolor{orange!10}
      Restormer~\cite{zamir2022restormer} & 26.11 & 282.2 & 39.16/0.986 & 37.11/0.985 & 36.36/0.977 & 35.00/0.971 & 38.42/0.978 &  37.21/0.979\\
      \rowcolor{gray!20}
      \textbf{TSANet-m (Ours)} & \textbf{8.74} & \textbf{81.94} & \textbf{39.24/0.986} & 37.25/\textbf{0.986} & \textbf{36.75/0.982} & 35.60/\textbf{0.976} & \textbf{38.93/0.979} & \textbf{37.55/0.982} \\
      \midrule
      \rowcolor{orange!10}
      DemosaicFormer~\cite{Xu_2024_CVPR} & 30.28 & 491.1 & 39.32/0.982 & \textbf{37.65}/0.982 & \textbf{37.64}/0.980 & 34.86/0.968 & \textbf{39.35/0.981} &   37.76/0.979\\
      \rowcolor{gray!15}
      \textbf{TSANet-l (Ours)} & \textbf{16.26} & \textbf{149.4} & \textbf{39.40/0.986} & 37.34/\textbf{0.986} & 37.07/\textbf{0.983} & \textbf{35.76/0.977} & 39.07/0.980 & \textbf{37.73/0.982} \\
      \bottomrule
    \end{tabular}
  }
  \label{tab:1}
  \vspace{-4mm}
\end{table*}

\begin{figure*}[t]
    \captionsetup[subfigure]{labelformat=empty}
    \centering
    \subfloat[Reference]{\includegraphics[width=0.19\textwidth]{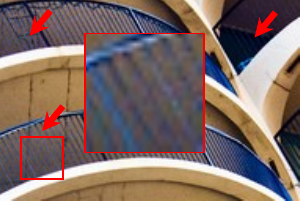}}
    \hfill
    \subfloat[DemosaicFormer~\cite{Xu_2024_CVPR}]{\includegraphics[width=0.19\textwidth]{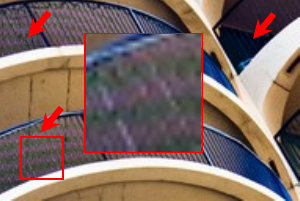}}
    \hfill
    \subfloat[NAFNet~\cite{chen2022simple}]{\includegraphics[width=0.19\textwidth]{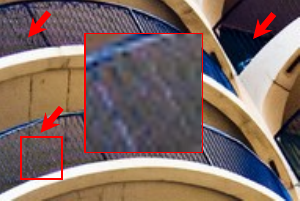}}
    \hfill
    \subfloat[SAGAN~\cite{sharif2021sagan}]{\includegraphics[width=0.19\textwidth]{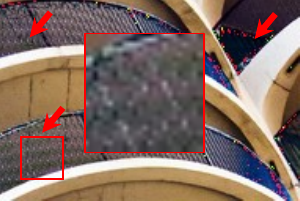}}
    \hfill
    \subfloat[SwinIR~\cite{liang2021swinir}]{\includegraphics[width=0.19\textwidth]{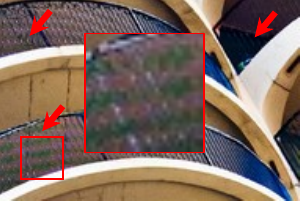}}
    
    \vfill
    
    \subfloat[UFormer~\cite{wang2022uformer}]{\includegraphics[width=0.19\textwidth]{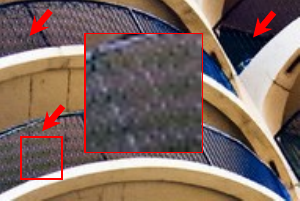}}
    \hfill
    \subfloat[PIPNet~\cite{a2021beyond}]{\includegraphics[width=0.19\textwidth]{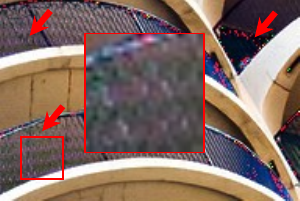}}
    \hfill
    \subfloat[Restormer~\cite{zamir2022restormer}]{\includegraphics[width=0.19\textwidth]{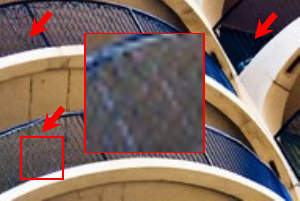}}
    \hfill
    \subfloat[CycleISP~\cite{zamir2020cycleisp}]{\includegraphics[width=0.19\textwidth]{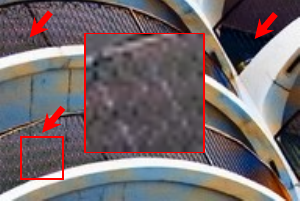}}
    \hfill
    \subfloat[TSANet-l (Ours)]{\includegraphics[width=0.19\textwidth]{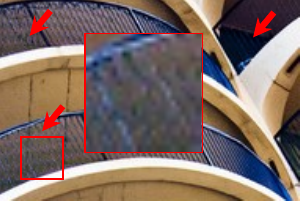}}

    \vfill

    \caption{Visualized results of all compared methods for Quad Bayer HybridEVS Demosaicing on synthesized image datasets. "Demosaicformer" is abbreviated as "DFormer". The comparison provides further validation of TSANet across various scenarios and confirms its effectiveness in restoring fine details, colors, and textures.}
    \label{figure6}
    \vspace{-3mm}
\end{figure*}

\begin{figure*}[t]
\captionsetup[subfigure]{labelformat=empty}
    \centering
    \subfloat[Reference]{\includegraphics[width=0.19\textwidth]{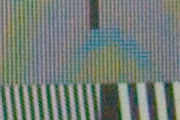}}
    \hfill
    \subfloat[DemosaicFormer~\cite{Xu_2024_CVPR}]{\includegraphics[width=0.19\textwidth]{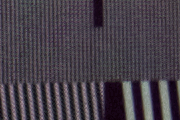}}
    \hfill
    \subfloat[NAFNet~\cite{chen2022simple}]{\includegraphics[width=0.19\textwidth]{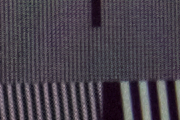}}
    \hfill
    \subfloat[SAGAN~\cite{sharif2021sagan}]{\includegraphics[width=0.19\textwidth]{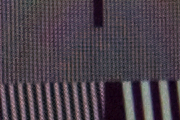}}
    \hfill
    \subfloat[SwinIR~\cite{liang2021swinir}]{\includegraphics[width=0.19\textwidth]{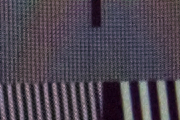}}
    
    \vfill

    \subfloat[UFormer~\cite{wang2022uformer}]{\includegraphics[width=0.19\textwidth]{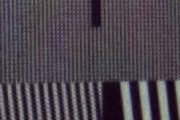}}
    \hfill
    \subfloat[Restormer~\cite{zamir2022restormer}]{\includegraphics[width=0.19\textwidth]{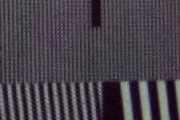}}
    \hfill
    \subfloat[PIPNet~\cite{a2021beyond}]{\includegraphics[width=0.19\textwidth]{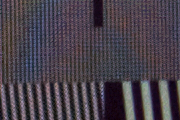}}
    \hfill
    \subfloat[CycleISP~\cite{zamir2020cycleisp}]{\includegraphics[width=0.19\textwidth]{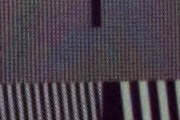}}
    \hfill
    \subfloat[TSANet-l (Ours)]{\includegraphics[width=0.19\textwidth]{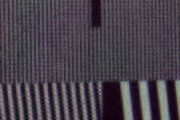}}

    \caption{Visualized results of real data. Reference is restored by using a gradient-based demosaicing method~\cite{RainbowJohnnyJohnny2024}. Our TSANet produces fewer moiré artifacts on the optical resolution test board with less parameters compared with top-rated methods.}
    \label{figure8}
    \vspace{-3mm}
\end{figure*}

\begin{figure*}[t]
\captionsetup[subfigure]{labelformat=empty}
    \centering
    \subfloat[Reference]{
    \includegraphics[width=0.13\textwidth]{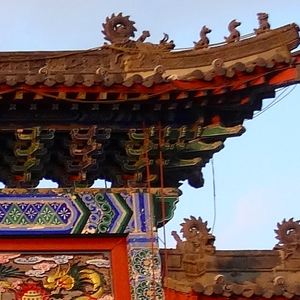}
    
    }
    \hfill
    \subfloat[Case 1]{\includegraphics[width=0.13\textwidth]{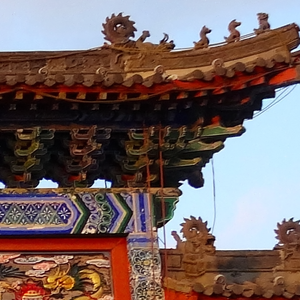}}
    \hfill
    \subfloat[Case 2]{\includegraphics[width=0.13\textwidth]{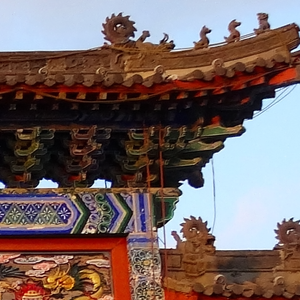}}
    \hfill
    \subfloat[Case 3]{\includegraphics[width=0.13\textwidth]{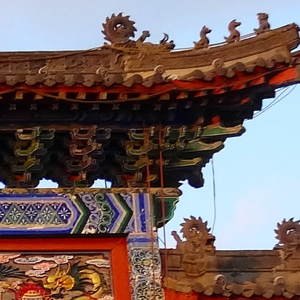}}
    \hfill
    \subfloat[Case 4]{\includegraphics[width=0.13\textwidth]{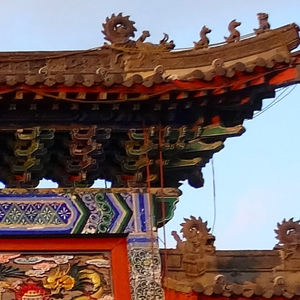}}
    \hfill
    \subfloat[Case 5]{\includegraphics[width=0.13\textwidth]{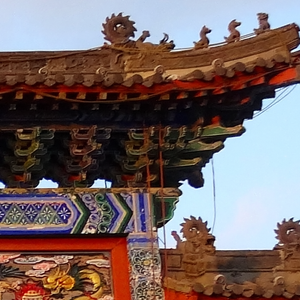}}
    \hfill
    \subfloat[Case 6]{\includegraphics[width=0.13\textwidth]{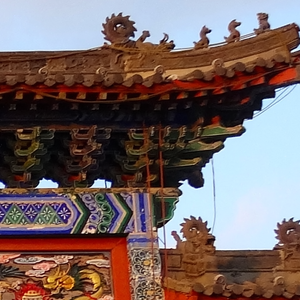}}
    
    \vfill
    
    \subfloat{\includegraphics[height=0.13\textwidth]{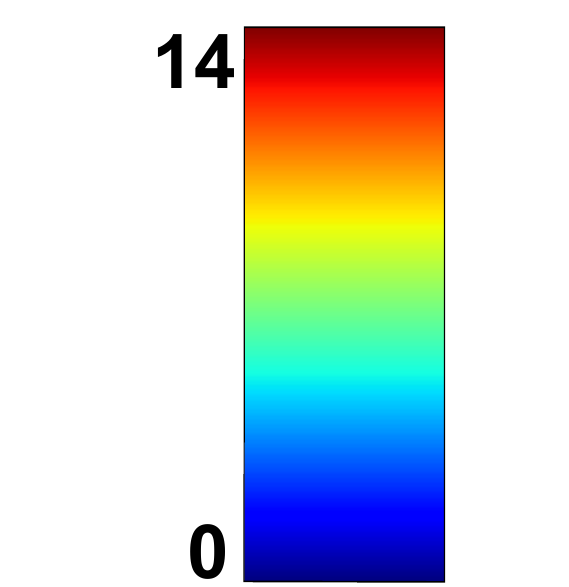}}
    \hfill
    \subfloat{\includegraphics[width=0.13\textwidth]{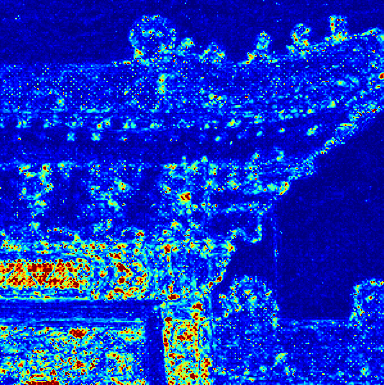}}
    \hfill
    \subfloat{\includegraphics[width=0.13\textwidth]{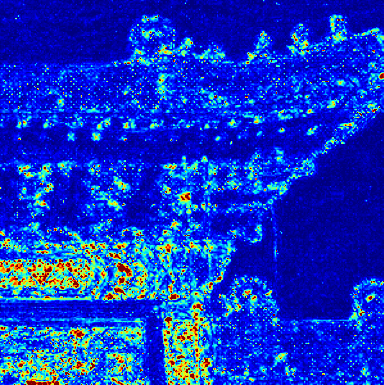}}
    \hfill
    \subfloat{\includegraphics[width=0.13\textwidth]{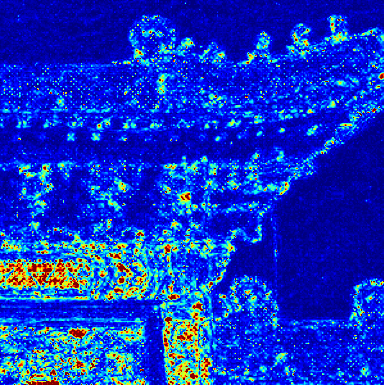}}
    \hfill
    \subfloat{\includegraphics[width=0.13\textwidth]{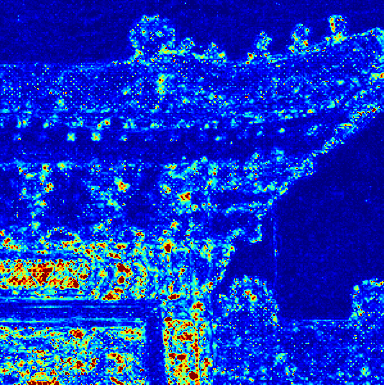}}
    \hfill
    \subfloat{\includegraphics[width=0.13\textwidth]{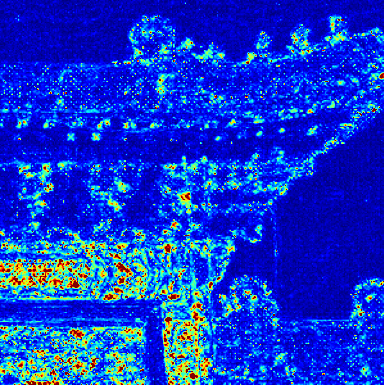}}
    \hfill
    \subfloat{\includegraphics[width=0.13\textwidth]{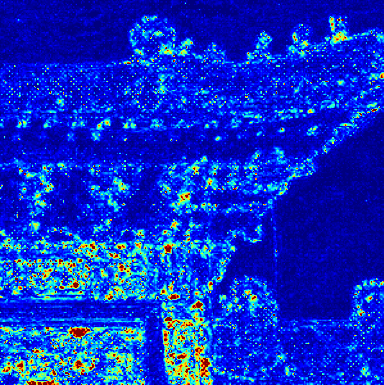}}
    
    \caption{Visualization of ablation study. We display the visual comparison results with difference maps of our ablation studies, which separately validate various components.}
    \label{figure9}
    \vspace{-4mm}
\end{figure*}

\paragraph{Datasets}
Both real data and synthesized data are used in our experiments. We train all the models with the dataset from Mobile Intelligent Photography \& Imaging (MIPI) Workshop 2024 Demosaic for Hybridevs Camera challenge track, which contains 800 pairs of Quad Bayer and RGB images with 2K resolution. The official public test set of MIPI dataset contains 26 pairs. Both the training and testing data include synthesized Gaussian noise and defect pixels~\cite{wu2024mipi}. We simulate HybridEVS pattern test cases from five image datasets, including Kodak~\cite{10.1145/1290082.1290117}, Urban100~\cite{Cordts_2016_CVPR}, BSD100~\cite{937655}, and Wed~\cite{7752930} (first 100 images), which includes most common scenes in daily life. Additionally, we assess dynamic performance using two video datasets: REDS~\cite{nah2019ntire} and Vid4~\cite{liu2011bayesian}. We re-sample Quad Bayer images from RGB images and simulate event pixels by setting them to 0 to generate the input. Real data is acquired utilizing a HybridEVS camera of 2448 × 3246 resolution. The influence of noise is minimized by increasing exposure time to validate the performance of demosaicing. Auto white balance is employed for color adjustment, and reference images shown in Fig. \ref{figure8} are obtained by using a gradient-based demosaicing method~\cite{RainbowJohnnyJohnny2024} to demonstrate the differance between classic and learning-based demosaicing methods.

\subsection{Comparison to State-of-the-Arts}
\vspace{-1mm}

\begin{table}[ht]
\caption{Quantitative comparison on video datasets REDS and Vid4. Our TSANet achieves state-of-the-art results across all model sizes.}
  \centering
  \small
  \setlength{\tabcolsep}{1mm}{
    \begin{tabular}{l|cc|c}
      \toprule
      \multirow{2}{*}{Methods} &REDS & Vid4 & \multirow{2}{*}{Average} \\
      \cline{2-3}
       &\multicolumn{2}{c|}{PSNR / SSIM} & \\
      \midrule
      \rowcolor{green!5}
      CycleISP~\cite{zamir2020cycleisp} & 32.96/0.975 & 30.46/0.964 & 31.71/0.970 \\
      \rowcolor{green!5}
      PIPNet~\cite{a2021beyond} & 36.19/0.981 & 32.20/0.964 & 34.20/0.972\\
      \rowcolor{green!5}
      SwinIR~\cite{liang2021swinir} & 41.38/0.994 & 34.57/0.977 & 37.98/0.986\\
      \rowcolor{green!5}
      UFormer~\cite{wang2022uformer} & 41.95/0.995 & 35.10/0.980 & 38.53/0.988 \\
      \rowcolor{gray!20}
      \textbf{TSANet-s (Ours)} & \textbf{41.94/0.989} & \textbf{35.15/0.980} & \textbf{38.55/0.984} \\
      \midrule
      \rowcolor{green!5}
      SAGAN~\cite{sharif2021sagan} & 38.13/0.984 & 32.16/0.963 & 35.14/0.974 \\
      \rowcolor{green!5}
      NAFNet~\cite{chen2022simple} & 42.76/0.996 & 35.48/0.981 & 39.12/0.988 \\
      \rowcolor{orange!10}
      Restormer~\cite{zamir2022restormer} & 41.91/0.990 & 35.08/0.980 &  38.49/0.985\\
      \rowcolor{gray!20}
      \textbf{TSANet-m (Ours)} & \textbf{42.87/0.996} & \textbf{35.53/0.982} & \textbf{39.20/0.989} \\
      \midrule
      \rowcolor{orange!10}
      DemosaicFormer~\cite{Xu_2024_CVPR} & 42.45/0.991 & \textbf{36.01}/0.979 &  39.23/0.985\\
      \rowcolor{gray!15}
      \textbf{TSANet-l (Ours)} & \textbf{43.00/0.996} & 35.64/\textbf{0.982} & \textbf{39.32/0.989} \\
      \bottomrule
    \end{tabular}
  }
  
  \vspace{-2mm}
  \label{tab:2}
\end{table}

\paragraph{Quantitative Comparison}
We compare our proposed TSANet with several state-of-the-art methods, including three dedicated demosaicing methods PIPNet~\cite{a2021beyond} and SAGAN~\cite{sharif2021sagan} based on Conlovlution and DemosaicFormer~\cite{Xu_2024_CVPR} based on Transformer, five image restoration methods, Restormer~\cite{zamir2022restormer}, SwinIR~\cite{liu2021swin} and UFormer~\cite{wang2022uformer} based on Transformer, NAFNet~\cite{chen2022simple} and CycleISP~\cite{zamir2020cycleisp} based on Convolution. We choose Peak Signal-to-Noise Ratio (PSNR) and Structural Similarity Index Measure (SSIM) scores as restoration quality metrics, while parameters and FLOPs as computational complexity metrics, to illustrate models' performance. Higher PSNR values indicate better reconstruction quality. SSIM values range between 0 (no similarity) and 1 (identical images), with higher values indicating better perceptual quality.

Table \ref{tab:1} and \ref{tab:2} shows the quantitative comparison results on all image and video datasets. It is worth noting that due to its large model size, Restormer and DemosaicFormer couldn't be tested on the 2k MIPI dataset, so we partitioned input images into 700x700 patches for inference and recombined the results. In particular, the proposed TSANet achieves state-of-the-art performance in different complexity levels across seven diverse test datasets, surpassing the previous SOTA method DemosaicFormer by 0.004 in SSIM, while reducing parameters and computations by 1.86x and 3.29x respectively. Additionally, on seven synthetic datasets, TSANet-l achieves the best PSNR/SSIM results in three and second-best results in four. Furthermore, smaller versions of TSANet-s and TSANet-m also achieve the best results on corresponding complexity ranges. Experimental comparisons validate the effectiveness of our approach, ensuring restoration performance while drastically reducing computational resources, introducing a model friendly to edge devices with limited computational resources.

\begin{table}[ht]
\caption{Quantitative results of ablation study. We validated the impact of the QCSA, SPA, RVSS, FFM modules and the two-step training strategy on TSANet-s.}
  \centering
  \small
  \setlength{\tabcolsep}{1mm}{
    \begin{tabular}{l|cccc|ccc }
      \toprule
      \multirow{2}{*}{Case}&\multicolumn{4}{c|}{Modules} &\multirow{2}{*}{\makecell{PARAMs\\(M)}}&\multirow{2}{*}{\makecell{FLOPs\\(G)}} & MIPI   \\
      \cmidrule(r){2-5}
      &QCSA& SPA&RVSS&FFM&&& PSNR/SSIM   \\
      \midrule
      1&&               &$\checkmark$ & $\checkmark$ & 3.70  & 32.0  & 38.41/0.977       \\ 
      2&$\checkmark$ &               &$\checkmark$ & $\checkmark$  &3.64  &32.5  &   38.42/0.978 \\ 
      3&& $\checkmark$  & $\checkmark$& $\checkmark$ & 4.06 & 36.9 &  38.45/0.978\\ 
      \midrule
      4&$\checkmark$ & $\checkmark$  &             & $\checkmark$ &6.42 & 59.1 & 38.49/0.978\\  
      5&$\checkmark$ & $\checkmark$  &$\checkmark$ &  & 3.99 & 36.2 & 38.44/0.978 \\ 
      \midrule
      6&$\checkmark$ & $\checkmark$  &$\checkmark$ & $\checkmark$  &4.00 & 37.4 & 38.47/0.978 \\ 
      \bottomrule
    \end{tabular}
  }
  \label{table2}
  \vspace{-4mm}
\end{table}
\paragraph{Visualization Comparison}
Fig. \ref{figure5}, \ref{figure6} and \ref{figure8} respectively depict the visual comparison results of TSANet compared with other models on synthesized image datasets, video datasets, and real HybridEVS data. Benefiting from our local-global feature extraction structure, our model exhibits closer visual similarity to ground truth in image details, particularly in subtle color variations, resulting in more vivid colors compared to other methods. While others suffer from color loss due to the Quad Bayer pattern and event pixels, leading to further degradation of color information in fine textures and even causing severe pixel errors (see Fig. \ref{figure5}). On the real HybridEVS data shown in Fig.\ref{figure8}, our TSANet achieved excellent restoration results with fewer moiré artifacts. Although top-rated methods produce similar outputs, TSANet offers the advantage of significantly fewer model parameters. Overall, the perceptual visual results confirm the effectiveness of our approach.

\subsection{Ablation Study}
\paragraph{Modules}
Our ablation study on the MIPI dataset validates the efficacy of the proposed modules (Table \ref{table2}). The cross-attention modules, QCSA and SPA, individually improve PSNR by 0.01 dB and 0.04 dB, respectively. Their combination yields a 0.06 dB gain with only a marginal 8\% increase in parameters. This confirms that explicitly encoding the position of event pixels via these modules provides critical guidance for the inpainting task. Furthermore, the RVSS module demonstrates a highly favorable efficiency-performance trade-off: replacing attention mechanisms with RVSS reduces parameters by a substantial 38\% at the cost of a minor 0.02 dB PSNR drop, highlighting the effectiveness of state space models. The qualitative impact of these modules on detail and texture restoration is visually demonstrated in Fig. \ref{figure9}.

\begin{table}[ht]
\caption{Parameters and FLOPs in each stage. }
  \centering
  \small
  \setlength{\tabcolsep}{0.8mm}{
    \begin{tabular}{c |c c c |c c c}
      \toprule
      \multirow{2}{*}{Method} &\multicolumn{3}{c|}{Params (M)} &\multicolumn{3}{c}{FLOPs (G)}\\
        &Q2Q&Q2R&Total &Q2Q&Q2R&Total\\
      \hline
      TSANet-s &1.49 &2.51&4.00&12.58 &24.86&37.44 \\
      TSANet-m &2.16 &6.58&8.74&17.73 &64.21&81.94 \\
      TSANet-l &2.16 &14.10&16.26&17.73 &131.63&149.36 \\
      \bottomrule
    \end{tabular}
  }
  \vspace{-1mm}
  \label{tab:details}
\end{table}
\begin{figure}[t]
\captionsetup[subfigure]{labelformat=empty}
    \centering
    \subfloat[GT]{\includegraphics[width=0.11\textwidth]{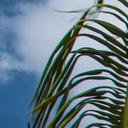}}
    \hfill
    \subfloat[TSANet-s]{\includegraphics[width=0.11\textwidth]{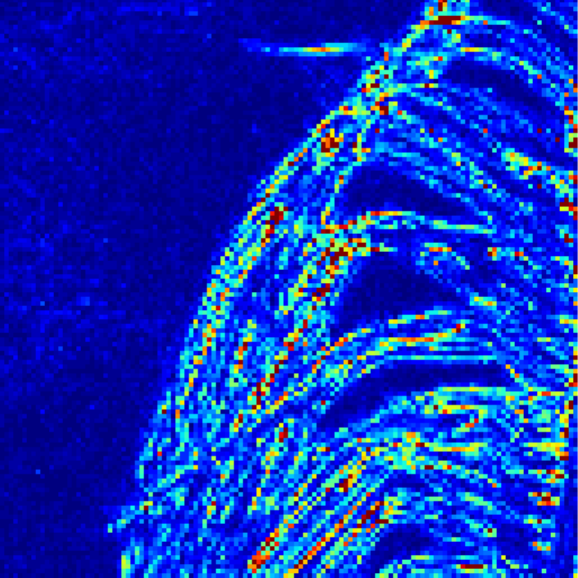}}
    \hfill
    \subfloat[TSANet-m]{\includegraphics[width=0.11\textwidth]{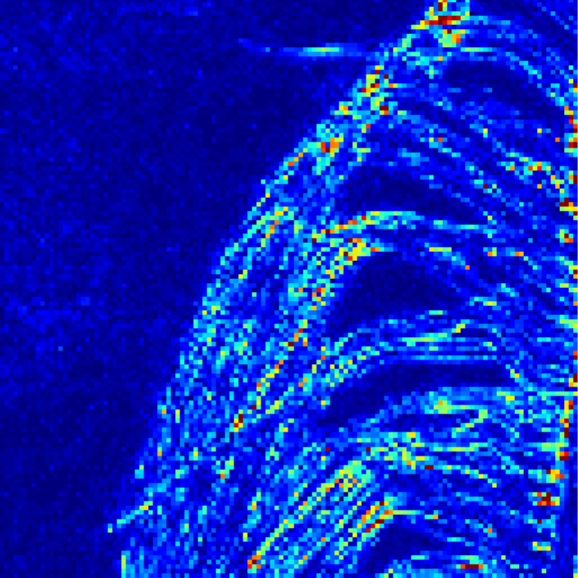}}
    \hfill
    \subfloat[TSANet-l]{\includegraphics[width=0.11\textwidth]{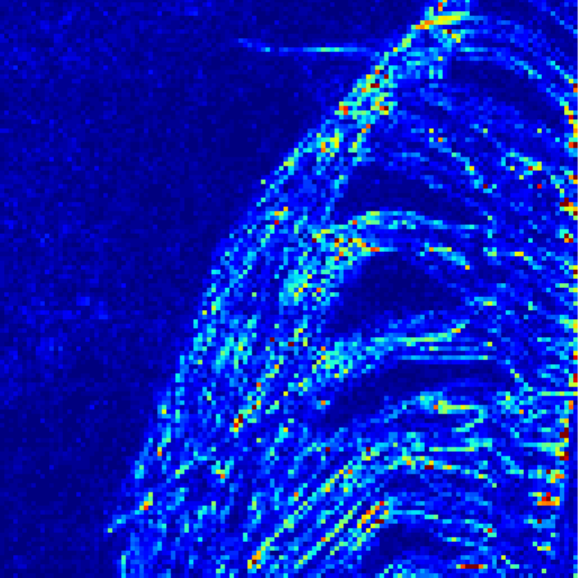}}
    \caption{Ablation study on model complexity. We compare three model scales: TSANet-s, -m, and -l. The difference maps reveal that larger models produce smaller reconstruction errors in regions with intricate details.}
    \label{figure10}
  \vspace{-4mm}
\end{figure}
\begin{table}[ht]
\caption{Ablation Study on the Two-Stage Strategy. TST means Two-stage training. }
  \centering
  \small
  \setlength{\tabcolsep}{0.5mm}{
    \begin{tabular}{c c c c c}
      \toprule
      Model & Params (M) & FLOPs (G) & TST & PSNR/SSIM \\
      \hline
      PIPNet        & 3.46      & 68.8      & N/A           & 33.73/0.950\\
      Q2Q + PIPNet  & 4.95      & 81.4      & $\checkmark$  & 35.33/0.963\\
      \hline
      \multirow{2}{*}{TSANet (Q2Q + Q2R)}  & \multirow{2}{*}{4.00}       & \multirow{2}{*}{37.4}      & $\times$  & 38.34/0.978\\
                       &        &      & $\checkmark$  & 38.47/0.978\\
      \bottomrule
    \end{tabular}
  }
  
    \vspace{-3mm}
  \label{tab:tst}
\end{table}

\begin{figure}[]
\captionsetup[subfigure]{labelformat=empty}
    \centering
    \subfloat[GT]{\includegraphics[width=0.15\textwidth]{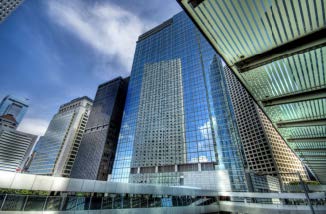}}
    \hfill
    \subfloat[TSANet-s w/o TST]{\includegraphics[width=0.15\textwidth]{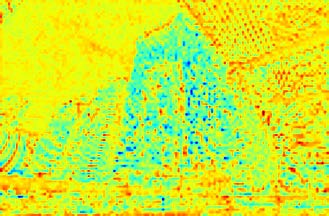}}
    \hfill
    \subfloat[TSANet-s with TST]{\includegraphics[width=0.15\textwidth]{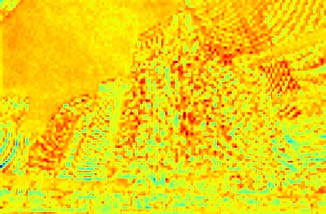}}
    
    \vfill
    
    \subfloat[GT]{\includegraphics[width=0.15\textwidth]{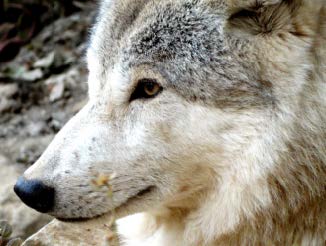}}
    \hfill
    \subfloat[PIPNet]{\includegraphics[width=0.15\textwidth]{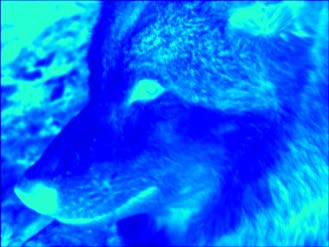}}
    \hfill
    \subfloat[Q2Q + PIPNet]{\includegraphics[width=0.15\textwidth]{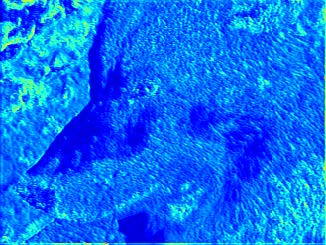}}
    
    \caption{Ablation Study on the Two-Step Design and Two-Stage Training (TST) Strategy. This figure visualizes the bottleneck feature maps from our TSANet and the PIPNet baseline~\cite{a2021beyond}, alongside the PIPNet model enhanced by our methods. The visualization demonstrates that incorporating the Two-Step Design and TST strategy significantly enhances the model's attention to fine-grained details within high-level features.}
    \label{figure11}
    \vspace{-4mm}
\end{figure}
We conducted an ablation study to validate our core two-stage design and two-step training (TST) strategy. This strategy features an asymmetric resource allocation principle (detailed in Table \ref{tab:details}), where the more complex Q2R demosaicing stage is intentionally assigned a higher-capacity network than the foundational Q2Q 'pixel-fix' stage.
The efficacy of this approach is confirmed in Table \ref{tab:tst}. Isolating the first component—task decomposition by integrating a Q2Q network into the PIPNet baseline yields a substantial 1.6 dB PSNR gain. This result validates our hypothesis that decomposing the complex mapping into simpler, sequential subtasks is highly effective. Furthermore, enabling our two-step training strategy contributes an additional 0.13 dB improvement without increasing inference cost, demonstrating the benefit of staged training for superior feature initialization.
Qualitatively, the feature maps in Fig. \ref{figure10} and \ref{figure11} reveal the impact of our approach. The visualizations show that our full model learns to focus on and restore fine-grained details that are overlooked by the baseline, providing visual confirmation of our strategy's efficacy in enhancing high-level feature representation.

\paragraph{Ablation on Two-stage training}
We then ablate the joint training strategy to determine how to best integrate the pretrained Q2Q and Q2R modules. As shown in Table \ref{tab:training_module}, we compare four approaches: direct concatenation (no joint training), freezing the Q2Q module, applying a dual-loss, and a single-loss end-to-end fine-tuning.
Both direct concatenation and freezing the Q2Q network lead to suboptimal performance, as the latter prevents full end-to-end optimization. More revealingly, the dual-loss strategy, despite preserving the quality of the intermediate Q2Q output, fails to produce the best final result. This exposes a critical insight: locally optimizing an intermediate stage can be counterproductive to achieving a globally optimal solution for the entire cascade.
Consequently, our model achieves the best performance with a single-loss, end-to-end training strategy. This confirms that allowing the entire network to adapt cohesively towards the final, singular objective is the most effective approach.
\begin{table}[ht]
\caption{Ablation Study on the joint training strategies. All experiments are conducted on TSANet-s and MIPI dataset.}
\centering
\resizebox{0.48\textwidth}{!}{
\setlength{\tabcolsep}{1mm} 
\begin{tabular}{c c c c}
  \toprule
  \multirow{2}{*}{Trainable modules} & \multirow{2}{*}{Loss} & \multicolumn{2}{c}{PSNR/SSIM} \\
                &              &Q2Q output                     &Final Output\\
  \hline
  None        & N/A                     & 42.61/0.995            & 38.12/0.975 \\
  Q2R           & Q2R loss              & 42.61/0.995            & 38.26/0.976 \\
  Q2Q + Q2R  & Q2Q + Q2R losses         & 38.30/0.992            &  38.33/0.976\\
  Q2Q + Q2R  & Q2R loss                 & 27.21/0.980            & \textbf{38.47/0.978}\\
  \bottomrule
\end{tabular}}

\label{tab:training_module}
\end{table}
\vspace{-4mm}
\section{Conclusion}

We have presented a novel lightweight two-stage structure network tailored for the HybridEVS demosaicing, which introduces task-specific sub-networks and a corresponding two-step training strategy. Specifically, we introduced two basic blocks strengthened by Residual Vision State Space (RVSS) to maintain low computational complexity while simultaneously addressing local position features and long-range dependencies. We further propose the two cross-modality attention mechanisms to effectively couple the arrangement information of Quad Bayer pattern and event points in the network, providing explicit prior encoding for global position information. To the best of our knowledge, this is the first work employing SSMs in a hybrid model for demosaicing tasks. Our approach is validated across multiple datasets and real HybridEVS data, offering a lightweight and mobile-friendly model for Quad Bayer HybridEVS Demosaicing.



\bibliographystyle{IEEEtran}
\bibliography{main}

\end{document}